\documentclass[conference]{IEEEtran}
\IEEEoverridecommandlockouts
\usepackage{cite}
\usepackage{amsmath,amssymb,amsfonts}
\usepackage{algorithmic}
\usepackage{graphicx}
\usepackage{textcomp}
\usepackage{xcolor}
\def\BibTeX{{\rm B\kern-.05em{\sc i\kern-.025em b}\kern-.08em
    T\kern-.1667em\lower.7ex\hbox{E}\kern-.125emX}}
\begin{document}

\title{Do Masked Autoencoders Improve Downhole Prediction? An Empirical Study on Real Well Drilling Data}

\author{
    \IEEEauthorblockN{1\textsuperscript{st} Aleksander Berezowski}
    \IEEEauthorblockA{\textit{Schulich School of Engineering} \\
    \textit{Department of Electrical and Software Engineering} \\
    \textit{University of Calgary}\\
    Calgary, Canada \\
    aleksander.berezowsk@ucalgary.ca}
\and
    \IEEEauthorblockN{2\textsuperscript{nd} Hassan Hassanzadeh}
    \IEEEauthorblockA{\textit{Schulich School of Engineering} \\
    \textit{Department of Chemical and Petroleum Engineering} \\
    \textit{University of Calgary}\\
    Calgary, Canada \\
    hhassanz@ucalgary.ca}
\and
    \IEEEauthorblockN{3\textsuperscript{rd} Gouri Ginde}
    \IEEEauthorblockA{\textit{Schulich School of Engineering} \\
    \textit{Department of Electrical and Software Engineering} \\
    \textit{University of Calgary}\\
    Calgary, Canada \\
    gouri.ginde@ucalgary.ca}
}

\maketitle

\begin{abstract}
Downhole drilling telemetry presents a fundamental labeling asymmetry: surface sensor data are generated continuously at 1~Hz, while labeled downhole measurements are costly, intermittent, and scarce. Current machine learning approaches for downhole metric prediction universally adopt fully supervised training from scratch, which is poorly suited to this data regime. We present the first empirical evaluation of masked autoencoder (MAE) pretraining for downhole drilling metric prediction. Using two publicly available Utah FORGE geothermal wells comprising approximately 3.5 million timesteps of multivariate drilling telemetry, we conduct a systematic full-factorial design space search across 72 MAE configurations and compare them against supervised LSTM and GRU baselines on the task of predicting Total Mud Volume. Results show that the best MAE configuration reduces test mean absolute error by 19.8\% relative to the supervised GRU baseline, while trailing the supervised LSTM baseline by 6.4\%. Analysis of design dimensions reveals that latent space width is the dominant architectural choice (Pearson $r = -0.59$ with test MAE), while masking ratio has negligible effect, an unexpected finding attributed to high temporal redundancy in 1~Hz drilling data. These results establish MAE pretraining as a viable paradigm for drilling analytics and identify the conditions under which it is most beneficial.
\end{abstract}


% ============================================================
\section{Introduction}
% ============================================================

Drilling operations generate dense multivariate telemetry at the surface, yet the downhole measurements that most directly reflect wellbore conditions, equivalent circulating density, bottom-hole pressure, string vibrations, and mud volume, are expensive to obtain and available only intermittently. Measurement-while-drilling (MWD) and logging-while-drilling (LWD) tools provide ground-truth downhole observations, but their deployment is costly and produces labeled datasets that are sparse relative to the volume of surface sensor data collected at 1~Hz throughout a drilling campaign. This asymmetry creates a challenging supervised learning scenario: models trained from scratch on labeled surface-to-downhole pairs must generalize from a small labeled pool while ignoring the vastly larger pool of unlabeled surface sequences.

The dominant machine learning paradigm in drilling analytics, as revealed by our systematic review of thirteen papers published between 2015 and 2025, is end-to-end supervised training on labeled pairs with no pretraining or representation learning component. Architectures range from artificial neural networks and gradient boosting to LSTM and GRU variants, but all share the assumption that sufficient labeled data are available to learn useful representations from scratch. This assumption is frequently violated in practice.

Masked autoencoders (MAEs), introduced in the computer vision domain by He et al.~\cite{He_Chen_Xie_Li_Dollar_Girshick_2022} and subsequently adapted to time-series data~\cite{li2023timaeselfsupervisedmaskedtime, tang2022mtsmaemaskedautoencodersmultivariate}, offer a compelling alternative. By training an encoder-decoder network to reconstruct randomly masked portions of an input sequence without supervision, MAE pretraining produces encoder representations that capture the underlying structure of the data. These representations can then be transferred to downstream regression tasks by appending a lightweight task header and training it on the available labeled data while keeping the encoder frozen. This two-stage transfer learning paradigm is data-efficient by design: pretraining exploits the abundant unlabeled surface sequences, and fine-tuning targets the scarce labeled downhole pairs.

Despite this natural alignment with the drilling analytics setting, no existing study has applied MAE pretraining, or any other form of self-supervised representation learning, to downhole metric prediction. We address that gap through three contributions. By leveraging masked autoencoder pretraining on abundant unlabeled surface telemetry, our work demonstrates a practical pathway toward data-efficient downhole prediction for AI-driven drilling optimization and energy-transition technologies.

First, we present the first application of MAE pretraining to drilling telemetry for downhole metric prediction. Second, we conduct a systematic design space exploration across 72 MAE configurations, varying encoder depth, latent space width, masking ratio, RNN cell type, and task header depth. Third, we provide a rigorous empirical comparison against supervised LSTM and GRU baselines trained on the same data under identical conditions, using publicly available Utah FORGE geothermal well data to ensure full reproducibility.

Our work makes the following concrete contributions to the drilling analytics community. We establish the first published MAE pretraining results on real-world multivariate drilling telemetry, providing a reproducible benchmark against which future self-supervised methods can be measured. Through a full-factorial search over 72 configurations, we identify latent space width as the dominant design choice ($r = -0.59$ with test MAE) and masking ratio as surprisingly inconsequential, findings that directly inform deployment decisions without requiring an expensive re-search. We provide the first empirical evidence that the high temporal redundancy of 1~Hz drilling telemetry suppresses the effectiveness of standard random masking strategies, motivating future work on structured masking tailored to this domain. Finally, all experiments are conducted on publicly available Utah FORGE geothermal data, ensuring that our baselines and MAE results can be extended or challenged by the community.

% ============================================================
\section{Related Work}
% ============================================================

This section reviews prior work across three threads that directly motivate the present study: (1) the surface and downhole metrics used in drilling analytics, (2) the machine learning methods currently applied to downhole metric prediction, and (3) the application of masked autoencoders to time-series data and their absence from the drilling domain.

\subsection{Surface and Downhole Metrics in Drilling Analytics}

We conducted a systematic mapping of thirteen papers published between 2015 and 2025 to identify the surface drilling measurements used as inputs to machine learning models and the downhole metrics predicted as outputs. Eight surface metrics and seven downhole metrics were identified, collectively defining the input-output scope of ML-based drilling analytics.

\begin{figure}[htbp]
\centerline{\includegraphics[width=0.5\textwidth,
    trim={.1cm .1cm .1cm 1cm}, clip]{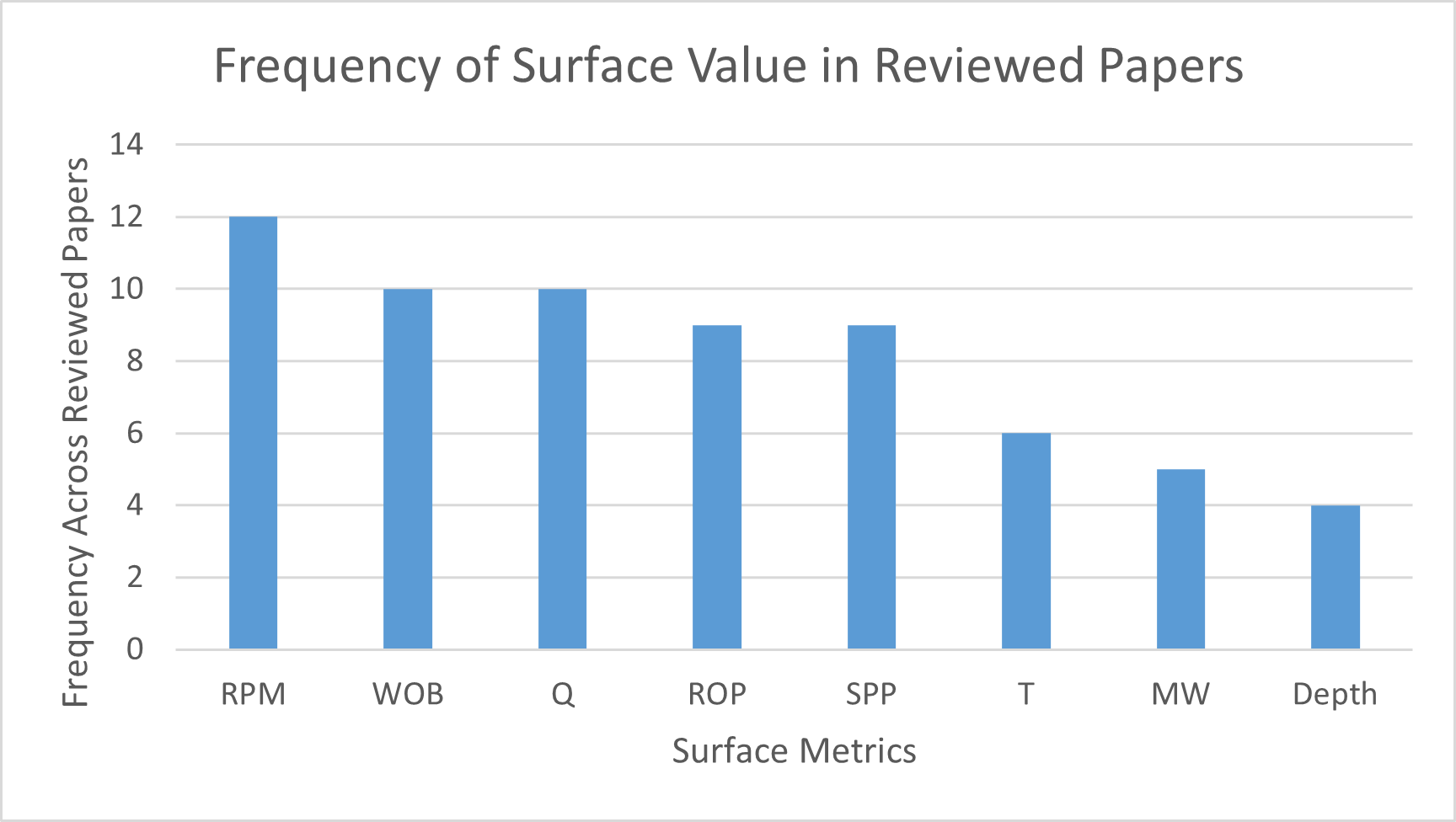}}
\caption{Frequency of surface metrics across the 13 reviewed papers. RPM, WOB, and flowrate (Q) are the most widely used inputs, appearing in 10 or more papers each. Metrics appearing in fewer than three papers were deemed insufficient for reliable generalization and excluded from our feature set. Readers should note the steep drop-off after SPP, which guided our feature selection to the top five most-reported channels.}
\label{surfaceChart}
\end{figure}

The eight surface metrics identified are rotations per minute (RPM), weight on bit (WOB), flowrate (Q), rate of penetration (ROP), standpipe pressure (SPP), surface torque (T), mud weight (MW), and depth. RPM was the most frequently used input, appearing in 12 of the 13 reviewed papers, followed by WOB and Q in 10 papers each, and ROP and SPP in 9 papers each. T appeared in 6 papers, MW in 5, and depth in 4. Metrics appearing in two or fewer papers, including outlet flow, conductivity, hookload, sand content, and back pressure, were considered insufficient for reliable generalization. The distribution of surface metrics across the reviewed papers is shown in Fig.~\ref{surfaceChart}.

\begin{figure}[htbp]
\centerline{\includegraphics[width=0.5\textwidth,
    trim={.1cm .1cm .1cm 1cm}, clip]{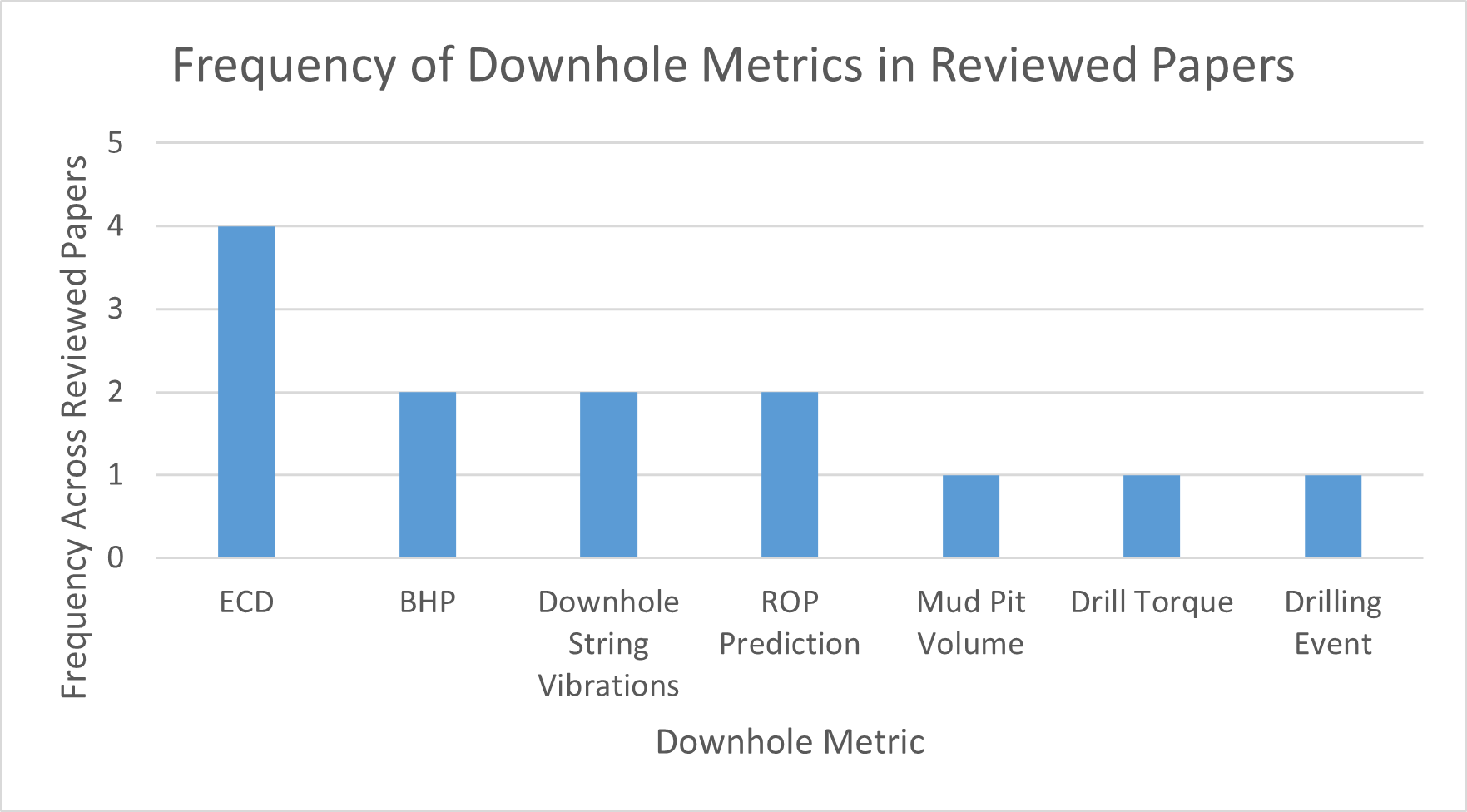}}
\caption{Frequency of predicted downhole metrics across the 13 reviewed papers. ECD dominates as the most-targeted output (4 papers), followed by BHP and downhole string vibrations (2 papers each). Total Mud Volume, the target in our study, appears in only one prior paper, highlighting the relatively underexplored nature of this prediction task and the opportunity for further investigation.}
\label{downholeChart}
\end{figure}

The seven predicted downhole metrics are equivalent circulating density (ECD), bottom-hole pressure (BHP), downhole string vibrations, rate of penetration (ROP) prediction, mud pit volume, drill torque, and drilling events. ECD was the most frequently predicted target, appearing in four papers, reflecting the importance of accurate ECD estimation for well control and the difficulty of achieving it with rule-based algorithms~\cite{EkechukwuGerald2024Empo, GamalHany2021MLMf, AbdelgawadKhaledZ.2019Nate, ZhaoWanchun2025Tdme}. BHP~\cite{ZhangCheng-Kai2023Bhpp, ZhangRui2023ANHT} and downhole string vibrations~\cite{SaadeldinRamy2022IMfP, SaadeldinRamy2023Ddvt} each appeared in two papers, while mud pit volume~\cite{ZhouYang2021Aohp}, drill torque~\cite{CAOWanpeng2025Dlat}, and drilling events~\cite{ZhaoJie2017MLTD} were each predicted by a single study. The distribution of predicted downhole metrics is shown in Fig.~\ref{downholeChart}.

\subsection{Machine Learning Methods for Downhole Metric Prediction}

Table~\ref{tab:ml-models-rw} summarizes the ML models used across the 13 reviewed papers. A total of 21 distinct methods were identified across the corpus, yet only 3 of those 21 methods appeared in more than one paper, reflecting the breadth of approaches currently being explored and the absence of an established standard.

\begin{table}[!h]
\renewcommand{\arraystretch}{1.2}
\caption{ML Methods Used for Downhole Metric Prediction}
\centering
\begin{tabular}{p{5.2cm}c}
\textbf{ML Method} & \textbf{Times Used} \\
\hline
Artificial Neural Network (ANN)                         & 4 \\
\hline
Adaptive Neuro-Fuzzy Inference System (ANFIS)           & 3 \\
\hline
Extreme Gradient Boosting (XGBoost)                     & 2 \\
\hline
Random Forest (RF)                                      & 1 \\
\hline
Backpropagation Neural Network (BPNN)                   & 1 \\
\hline
Convolutional Neural Network (CNN)                      & 1 \\
\hline
Gated Recurrent Unit (GRU)                              & 1 \\
\hline
CNN-GRU                                                 & 1 \\
\hline
Radial Basis Function network (RBF)                     & 1 \\
\hline
Fuzzy Network (FN)                                      & 1 \\
\hline
Support Vector Machine (SVM)                            & 1 \\
\hline
Long Short-Term Memory (LSTM)                           & 1 \\
\hline
LSTM with Domain-Adversarial NN (LSTM-DANN)             & 1 \\
\hline
SVR-BPNN-LSTM                                           & 1 \\
\hline
Deep Adversarial Neural Network (DARN)                  & 1 \\
\hline
Symbolic Aggregate Approximation (SAX)                  & 1 \\
\hline
Hierarchical Clustering                                 & 1 \\
\hline
Boosting                                                & 1 \\
\hline
Pruned Tree                                             & 1 \\
\hline
Random Forests                                          & 1 \\
\hline
Decision Tree                                           & 1 \\
\hline
\end{tabular}
\label{tab:ml-models-rw}
\end{table}

Neural networks of some form, including ANNs, LSTMs, CNNs, GRUs, and their hybrids, account for 14 of the 21 methods identified, underscoring their suitability for capturing the nonlinear temporal relationships in drilling data. Among these, ANN was the most frequently applied architecture (4 papers), reflecting its established role in the domain~\cite{GamalHany2021MLMf, AbdelgawadKhaledZ.2019Nate, ZhaoWanchun2025Tdme, SaadeldinRamy2022IMfP}. LSTM-based methods appeared in three distinct forms across the corpus: a standalone LSTM for ROP prediction~\cite{EncinasMauroA.2022Ddcf}, an LSTM combined with a domain-adversarial network for BHP transfer learning~\cite{ZhangRui2023ANHT}, and an SVR-BPNN-LSTM hybrid for mud pit volume prediction~\cite{ZhouYang2021Aohp}. GRU appeared once as a standalone model and once in a CNN-GRU hybrid, both for BHP prediction~\cite{ZhangCheng-Kai2023Bhpp}. Tree-based ensemble methods (Random Forest, XGBoost, Decision Tree, Boosting, Pruned Tree) appeared in four papers and were used primarily for ECD prediction~\cite{EkechukwuGerald2024Empo} and ROP~\cite{HegdeChiranth2015UTBa}.

Critically, all reviewed methods frame downhole metric prediction as a standard supervised learning problem: a model is trained from scratch on labeled surface-to-downhole pairs for a single target variable. None of the 13 reviewed papers employed autoencoder-based architectures, self-supervised pretraining, or foundation model paradigms of any kind. This represents a significant methodological gap, as such architectures are well-suited to settings where labeled downhole measurements are scarce but unlabeled surface data is abundant.

\subsection{The Research Gap: No Self-Supervised Learning in Drilling Analytics}

The absence of self-supervised or representation learning methods from the drilling analytics literature constitutes a clear and consequential gap. Three properties of the drilling data regime make this gap particularly striking. First, the labeling asymmetry is severe: surface sensor data are recorded continuously at 1~Hz throughout a drilling campaign, while labeled downhole measurements from MWD/LWD tools are intermittent, expensive, and often unavailable for the majority of a well's duration. Second, the unlabeled surface data are multivariate, temporally structured, and physically coherent, exactly the properties that self-supervised methods such as MAEs are designed to exploit. Third, the broader machine learning community has demonstrated that self-supervised pretraining substantially reduces the labeled data required to achieve strong downstream performance on analogous time-series tasks~\cite{li2023timaeselfsupervisedmaskedtime, tang2022mtsmaemaskedautoencodersmultivariate}, yet no drilling analytics paper has translated this insight to the downhole prediction problem.

The practical consequence of this gap is that every new well or formation requires training a model from scratch on whatever labeled data happen to be available, with no mechanism to leverage the far larger pool of unlabeled surface telemetry collected from the same or related wells. This is a missed opportunity: a pretrained encoder that captures the latent structure of drilling dynamics could be fine-tuned with only a small labeled set, reducing the cost and time required to deploy a prediction model in a new drilling context. Our work directly addresses this gap by providing the first empirical evaluation of MAE pretraining for downhole metric prediction, establishing both a methodology and a set of baselines against which future self-supervised drilling analytics methods can be measured.

\subsection{Masked Autoencoders for Time-Series Representation Learning}

Masked Autoencoder Foundation Models (MAEFMs) were introduced by He et al.~\cite{He_Chen_Xie_Li_Dollar_Girshick_2022} in the context of computer vision, where masking large fractions of image patches and training an encoder-decoder to reconstruct them yielded powerful, generalizable representations. The underlying autoencoder architecture traces to Rumelhart et al.~\cite{Rumelhart_Hinton_Williams_1986}, and the foundation model paradigm, in which a single model is pretrained on large unlabeled datasets and then fine-tuned on small task-specific datasets, is described by Schneider et al.~\cite{Schneider_Meske_Kuss_2024}.

Two studies establish the effectiveness of masked autoencoders on time-series data specifically. Li et al.~\cite{li2023timaeselfsupervisedmaskedtime} evaluated masked autoencoders across five benchmark datasets, including electricity transformer temperature, weather forecasting, exchange rate prediction, influenza forecasting, and the UCR time-series archive~\cite{UCRArchive2018}, finding that masked autoencoders consistently outperformed or matched other transformer-based architectures. Tang et al.~\cite{tang2022mtsmaemaskedautoencodersmultivariate} demonstrated that masked autoencoders are effective on multivariate time-series forecasting tasks, achieving performance comparable to or better than LSTMs and transformer models on electricity consumption, weather, and electricity transformer temperature benchmarks. Their results show that the architecture captures both temporal and cross-variable dependencies, a property of direct relevance to drilling telemetry, in which multiple correlated surface measurements (e.g., RPM, WOB, SPP, and flowrate) interact over time.

After MAEFM pretraining, the decoder is discarded, the encoder weights are frozen, and lightweight task-specific adapter layers are appended and trained on small labeled datasets~\cite{houlsby2019parameterefficienttransferlearningnlp}. This paradigm is directly applicable to oil and gas drilling, where surface sensor data are generated continuously and in large volumes, while labeled downhole measurements are expensive and sparse. Despite this alignment, the systematic mapping conducted as part of this study confirmed that no existing drilling analytics study has applied masked autoencoders or any other form of self-supervised representation learning to downhole metric prediction. The present paper addresses this gap by evaluating a masked autoencoder pretraining approach against supervised LSTM and GRU baselines on real multivariate well data.

% ============================================================
\section{Dataset and Feature Engineering}
% ============================================================

\subsection{Dataset Description}

The dataset used in this study comprises drilling telemetry from two wells drilled as part of the Utah Frontier Observatory for Research in Geothermal Energy (Utah FORGE) project, obtained from the United States Government's open data catalogue~\cite{Catalog}. Utah FORGE was selected for three reasons. First, its data are fully public and non-proprietary, making all experiments in our study completely reproducible by the broader community, a property that is rare in drilling analytics research, where most datasets are held under commercial confidentiality. Second, the project provides two wells drilled in the same geothermal formation with the same rig instrumentation, giving us a matched pair that enables balanced two-well pooling without confounding differences in sensor configuration or geological context. Third, the 1~Hz sampling rate and multi-channel instrumentation of the FORGE wells match the data regime assumed throughout our study, where abundant unlabeled surface sequences are available but labeled downhole measurements are scarce. Each well was instrumented at a sampling frequency of 1~Hz, and both wells contain considerably more sensor traces than those used here. Based on the surface metrics identified in the literature review conducted as part of this study, nine channels were selected from each well: Weight on Bit (WOB), Rotary RPM, Total Pump Output, Rate of Penetration (ROP), Standpipe Pressure (SPP), Rotary Torque, Hole Depth, Bit Depth, and Total Mud Volume. After filtering to active drilling intervals and combining both wells, the dataset comprises approximately 3.5 million timesteps in total.

\subsection{Feature Selection}

\begin{figure}[htbp]
\centerline{\includegraphics[width=0.4\textwidth,
    clip]{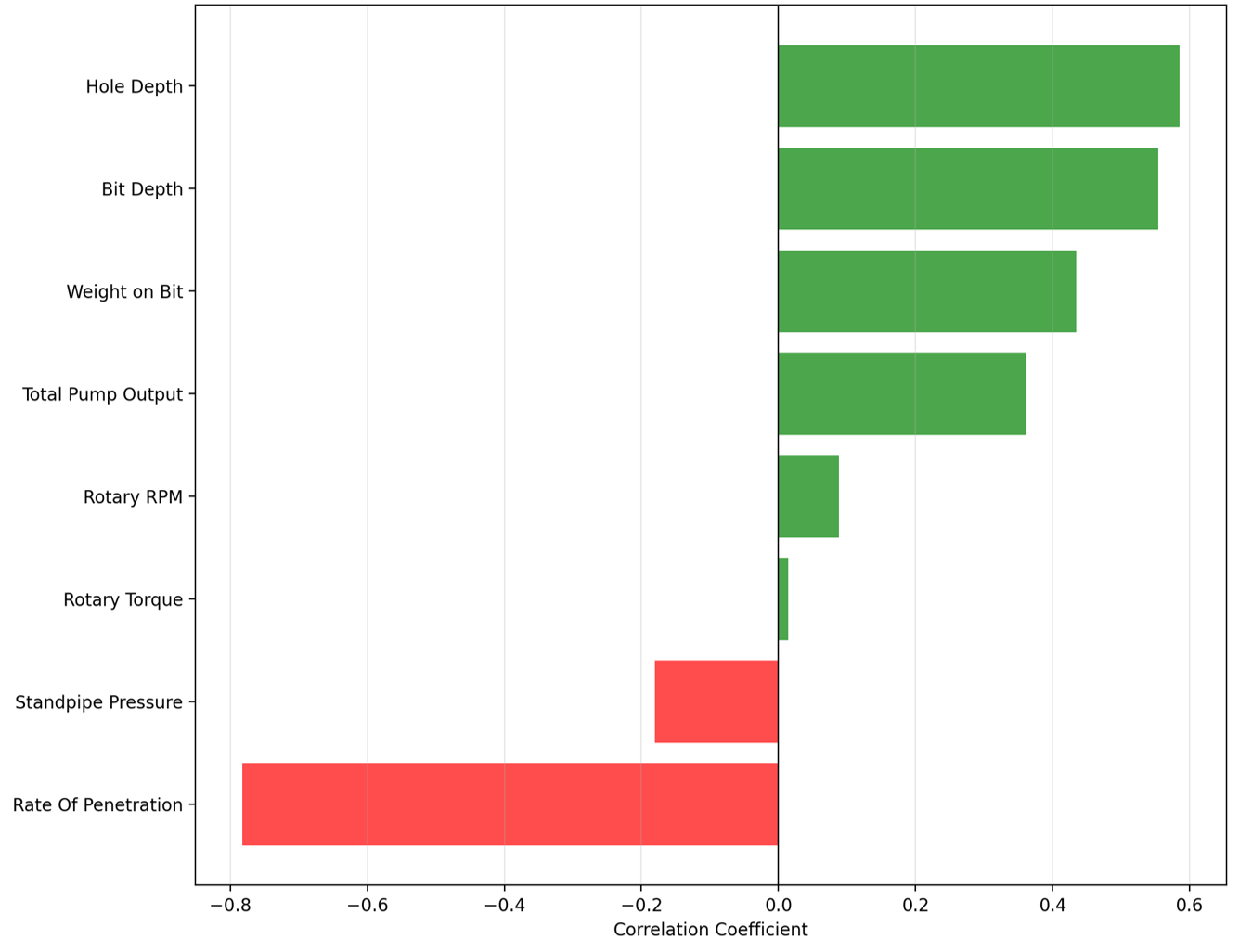}}
\caption{Pearson correlation coefficient between each input feature and the prediction target, Total Mud Volume. Green bars indicate positive correlation; red bars indicate negative correlation. Readers should note that Rate of Penetration has the largest absolute correlation ($r = -0.78$) despite being negative, confirming its strong but inverse relationship with mud volume. Features with $|r| < 0.2$ (Rotary RPM, Rotary Torque, Standpipe Pressure) were discarded as inputs. The key takeaway is that five channels carry meaningful linear signal for the prediction task while three do not, motivating the $600 \times 5$ input shape used throughout our study.}
\label{AllFeaturesCorrelationToMudVolume}
\end{figure}

\begin{figure}[htbp]
\centerline{\includegraphics[width=0.5\textwidth,
    clip]{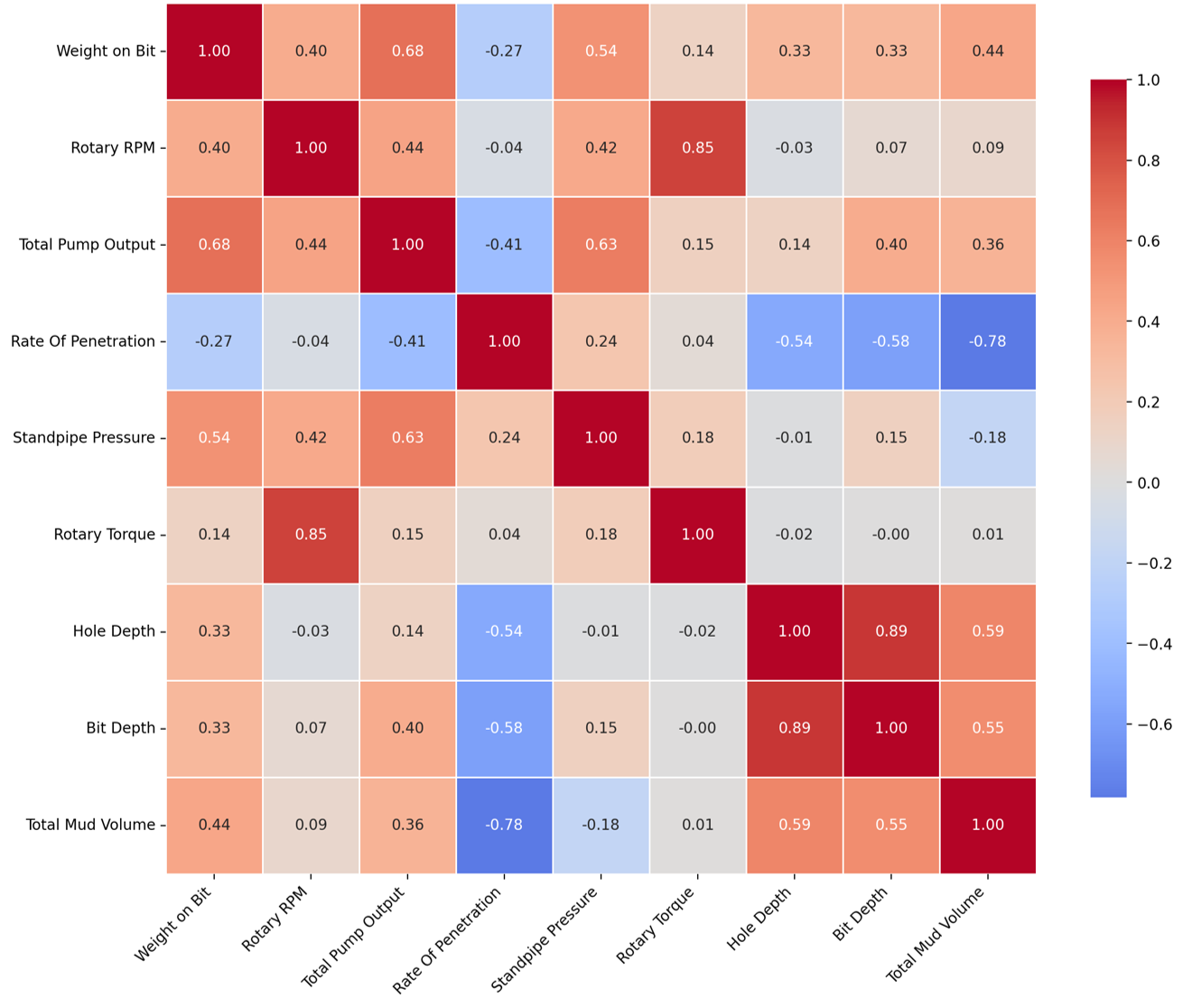}}
\caption{Pearson correlation matrix across all nine sensor channels. Strong red cells indicate high positive correlation; strong blue cells indicate high negative correlation. Readers should note two collinear pairs among the retained features: Hole Depth and Bit Depth ($r = 0.89$), expected because the bit tracks the hole during active drilling, and Rotary RPM and Rotary Torque ($r = 0.85$), arising from their shared mechanical coupling. Rate of Penetration shows strong negative correlation with the depth-related channels ($r \approx -0.55$), consistent with the physical relationship between faster penetration and shorter column depths. The key takeaway is that despite pairwise collinearity within the retained set, each of the five selected features contributes distinct signal to Total Mud Volume prediction.}
\label{AllFeaturesCorrMatrix}
\end{figure}

Total Mud Volume is used as the prediction target throughout this study. To select the input features from the remaining eight channels, Pearson correlation coefficients between each channel and Total Mud Volume were computed across all time-averaged windows. The results are shown in Fig.~\ref{AllFeaturesCorrelationToMudVolume}, and the full inter-feature correlation matrix is shown in Fig.~\ref{AllFeaturesCorrMatrix}.

Four features exhibit strong positive correlation with Total Mud Volume: Hole Depth ($r = 0.59$), Bit Depth ($r = 0.55$), WOB ($r = 0.44$), and Total Pump Output ($r = 0.36$). ROP shows the strongest relationship of any feature but in the negative direction ($r = -0.78$), reflecting the physical relationship between faster penetration and reduced mud column volume. The remaining three channels, Rotary RPM ($r = 0.09$), Rotary Torque ($r = 0.01$), and Standpipe Pressure ($r = -0.18$), exhibit weak linear relationships with the target and were discarded as input features.

The inter-feature correlation matrix reveals two noteworthy collinear pairs among the retained features. Hole Depth and Bit Depth are highly correlated ($r = 0.89$), which is expected since the bit tracks the hole during active drilling. Rotary RPM and Rotary Torque are also strongly correlated ($r = 0.85$), consistent with their shared mechanical coupling at the drillstring; both were excluded on the basis of low target correlation regardless. WOB and Total Pump Output show moderate correlation ($r = 0.68$), but both carry sufficient independent signal with the target to justify retention.

The five features selected as model inputs are therefore: WOB, ROP, Hole Depth, Bit Depth, and Total Pump Output. This selection is grounded both in the empirical correlations observed in this dataset and in the literature, where WOB, ROP, flowrate (Total Pump Output), and depth are among the most consistently reported surface metrics for ML-based downhole prediction.

\subsection{Drilling Activity Segmentation}

Raw drilling telemetry contains long intervals during which no active drilling occurs, including tripping, connections, and equipment standby periods. These idle intervals introduce uninformative signal and would corrupt both the self-supervised pretraining objective and the regression target. A multi-condition mask was therefore constructed to isolate contiguous active drilling segments before any further processing.

\subsubsection*{Stage 1: Timestep-level drilling indicator}

At each timestep $t$, a binary drilling indicator $m_t$ is computed by requiring three physical conditions to hold simultaneously:

\begin{equation}
    m_t = \left[\overline{HD}_{10000}' > 0\right]
    \wedge \left[HD_t = BD_t\right]
    \wedge \left[HD_t > 1000\right]
\label{eq:basemask}
\end{equation}

\noindent where $HD_t$ and $BD_t$ are Hole Depth and Bit Depth at timestep $t$, and $\overline{HD}_{10000}'$ is the first difference of the 10{,}000-sample rolling mean of Hole Depth. The first condition confirms that the well is deepening over a long horizon, ruling out flat or slowly drifting sensor readings during non-drilling activity. The second confirms the bit is at the bottom of the hole and not being tripped in or out. The third excludes the shallow surface phase, where drilling behavior differs substantially from the deeper intervals that make up the bulk of the dataset. Together, these three conditions produce a sparse binary signal that fires only when the rig is actively drilling.

\subsubsection*{Stage 2: Smoothing to identify sustained drilling regions}

Because the indicator $m_t$ can flicker at individual timesteps due to sensor noise or momentary pauses, it is smoothed in two passes to identify only sustained drilling regions. First, a 100-sample rolling mean is applied and thresholded at 0.3: a timestep is kept only if at least 30\% of the preceding 100 samples were flagged as drilling. This removes brief spurious firings. Second, a 20{,}000-sample rolling mean is applied to the result and thresholded at 0.6: a timestep is retained only if it falls within a window where the rig was drilling more than 60\% of the time across the surrounding ${\sim}5.5$~hours of data. This second pass selects large, coherent drilling blocks and discards everything outside them, including the extended idle periods between runs.

\subsubsection*{Stage 3: Grouping into contiguous segments}

Timesteps that survive both smoothing passes are retained; all others are set to NaN. The indices of retained timesteps are then scanned sequentially: if the gap between two consecutive retained indices exceeds 100 samples, the current run is closed and a new group begins. Each resulting group is a contiguous block of timesteps belonging to a single sustained drilling interval. Any NaN values remaining within a group's numeric channels are imputed using a 100-sample centered rolling mean, with backward-fill and forward-fill applied at the segment boundaries to ensure no NaN values propagate into the windowing step.

\subsection{Sliding Window Construction}

Each drilling segment is converted into overlapping fixed-length windows using a stride of one sample. The window length is set to $600$ timesteps, corresponding to 10~minutes of continuous drilling data at the 1~Hz sampling rate. This duration was chosen for three reasons. First, it captures a physically meaningful horizon of drilling dynamics: 10~minutes is long enough to encode trends in depth advancement, pressure buildup, and weight progression while being short enough that the window-mean of Total Mud Volume remains a well-defined local summary statistic that does not blur across distinct drilling regimes. Second, a 600-sample window is consistent with the short-to-medium context lengths used in recurrent sequence models for industrial time-series tasks, where windows of 256--1024 samples are typical~\cite{tang2022mtsmaemaskedautoencodersmultivariate, li2023timaeselfsupervisedmaskedtime}. Third, at stride 1 the 600-sample window generates a large number of overlapping sequences from each drilling segment, providing the pool of unlabeled pretraining examples required for effective self-supervised MAE training. The resulting input tensor shape of $(600, 5)$ thus reflects a deliberate trade-off between temporal context, prediction target stability, and pretraining data volume.

Formally, for a segment of length $T$, windows $\mathbf{X}^{(i)} \in \mathbb{R}^{600 \times F}$ are extracted as:

\begin{equation}
    \mathbf{X}^{(i)} = \mathbf{S}[i \,:\, i + 600], \quad
    i = 0, 1, \ldots, T - 600
\end{equation}

\noindent where $\mathbf{S} \in \mathbb{R}^{T \times F}$ is the segment matrix and $F$ is the number of selected features. The stride-1 sliding window produces a large pool of overlapping sequences, which provides the abundance of unlabeled data required for effective self-supervised MAE pretraining.

\subsection{Normalization}

All channels are normalized to the range $[0, 1]$ using min-max scaling. Critically, the global minimum and maximum for each channel are computed across the union of all drilling segments before any windows are extracted:

\begin{equation}
    \hat{x} = \frac{x - x_{\min}}{x_{\max} - x_{\min}}
\end{equation}

Computing statistics at the segment level rather than the window level, and doing so prior to the train/test split, ensures two properties. First, the scale of each channel is consistent across all windows from both wells, preventing the model from exploiting inter-well amplitude differences as a shortcut. Second, because the global statistics are computed over complete segments rather than individual windows, there is no leakage of window-level target information into the normalization parameters.

\subsection{Feature Engineering and Target Definition}

Following feature selection, the model operates on $F = 5$ input channels: WOB, ROP, Total Pump Output, Hole Depth, and Bit Depth. The prediction target, Total Mud Volume, is extracted from each window and then removed from the input tensor so the encoder cannot trivially copy it. Specifically, the target for window $\mathbf{X}^{(i)}$ is defined as the temporal mean of Total Mud Volume over the 600-timestep horizon:

\begin{equation}
    y^{(i)} = \frac{1}{600} \sum_{t=1}^{600} x^{(i)}_{t,\,\text{MV}}
\end{equation}

\noindent where $x^{(i)}_{t,\,\text{MV}}$ denotes the normalized Total Mud Volume at timestep $t$ in window $i$. Window-mean aggregation is appropriate for this prediction horizon for two reasons. First, mud volume changes slowly relative to the 1~Hz sampling rate; the within-window variance is low compared to between-window variance, so the mean is a stable and representative summary. Second, predicting a scalar summary statistic rather than a full future sequence simplifies the task header architecture, keeping the regression problem tractable and directly comparable to the supervised baselines.

After target extraction and column removal, each window is a tensor of shape $(600, 5)$ and its associated label is a single scalar in $[0, 1]$.

\subsection{Dataset Balancing and Splits}

The two wells differ in total length, which would cause the combined window pool to be dominated by the longer well if windows were pooled without correction. To prevent well-specific bias in both the pretraining and fine-tuning stages, the window lists from each well are independently shuffled with a fixed random seed and then truncated to the length of the shorter well before concatenation. This ensures both wells contribute equally to the final pool regardless of their original lengths.

The balanced pool is then split into training and test sets using an 80/20 ratio with stratified random sampling, yielding reproducible splits. The test set is held out entirely and used only for final evaluation; no hyperparameter decisions are made on the basis of test set performance.

To make the full design space search computationally feasible, a random subset of 20\% of the balanced window pool is used throughout all experiments. This subsampling is applied after balancing and before splitting, so both the train and test sets are drawn from the same proportionally reduced pool and the 80/20 ratio is preserved. All reported results, including the baseline comparisons, are obtained under this consistent 20\% subset condition.

% ============================================================
\section{Methodology}
% ============================================================

This section describes all models evaluated in this study and the procedures used to train and compare them. Three model families are considered: a supervised LSTM baseline, a supervised GRU baseline, and a family of masked autoencoder (MAE) models trained via a two-stage transfer learning pipeline. A systematic design space search is used to identify the best-performing MAE configuration, which is then compared against both baselines on a held-out test set.

\subsection{Baseline Supervised Models}

Two fully supervised recurrent baselines are trained to establish the performance level achievable by the current state of practice for downhole metric prediction. LSTM and GRU were selected as the baseline architectures for three reasons. First, our literature review identified them as the dominant recurrent architectures applied to sequential drilling telemetry: LSTM appeared in three distinct forms across the 13 reviewed papers, and GRU appeared both standalone and in a CNN-GRU hybrid~\cite{ZhangCheng-Kai2023Bhpp, ZhangRui2023ANHT, ZhouYang2021Aohp, EncinasMauroA.2022Ddcf}. Second, both architectures are specifically designed to model long-range temporal dependencies via gating mechanisms, which makes them well-suited to the autocorrelated, slow-varying nature of drilling telemetry~\cite{HochreiterSepp1997Lstm, ChoKyunghyun2014GRU}. Third, both cells are the recurrent building blocks of our own MAE models, so comparing their fully supervised versions against the MAE pretraining pipeline provides the most direct and controlled test of what pretraining adds. Both baselines follow the same architecture and training procedure, differing only in the choice of recurrent cell.

Each baseline consists of four layers. The first recurrent layer processes the full input sequence of shape $(600, 5)$ with $\mathtt{return\_sequences=True}$, passing its hidden state at every timestep to a Dropout layer. The second recurrent layer then consumes the dropout-regularized sequence, collapsing the temporal dimension to produce a single fixed-length context vector. A final Dense layer with a single linear output unit maps this vector to the scalar mud volume prediction. Both recurrent layers use 64 hidden units, a width consistent with the scale used in comparable RNN-based drilling prediction studies~\cite{ZhangCheng-Kai2023Bhpp, ZhouYang2021Aohp} and sufficient to capture the dependencies among the five input channels without over-parameterizing for the available data volume. The LSTM baseline uses LSTM cells at both recurrent positions; the GRU baseline substitutes GRU cells at both positions. All other settings are identical.

Both baselines are compiled with the Adam optimizer~\cite{KingmaDiederikP2014Adam} at a learning rate of $0.001$ and trained to minimize mean absolute error (MAE). Adam was chosen over SGD and RMSProp because its adaptive per-parameter learning rates accelerate convergence on the heterogeneous feature scales present in normalized drilling telemetry, and a learning rate of $0.001$ is the widely recommended default for Adam on regression tasks~\cite{KingmaDiederikP2014Adam}. A batch size of 64 balances stochastic gradient noise and memory efficiency and is consistent with the batch sizes used in related time-series regression work~\cite{tang2022mtsmaemaskedautoencodersmultivariate}. Training runs for up to 15 epochs with early stopping (patience 5, monitoring validation loss) restoring the best weights before evaluation. A batch size of 64 is used throughout. Table~\ref{tab:baseline_config} summarizes the shared configuration.

\begin{table}[!h]
\renewcommand{\arraystretch}{1.2}
\caption{Baseline Model Configuration (LSTM and GRU)}
\centering
\begin{tabular}{ll}
\textbf{Hyperparameter} & \textbf{Value} \\
\hline
Hidden units (both layers)  & 64 \\
Dropout rate                & 0.3 \\
Optimiser                   & Adam \\
Learning rate               & 0.001 \\
Loss function               & MAE \\
Batch size                  & 64 \\
Max epochs                  & 15 \\
Early stopping patience     & 5 \\
\hline
\end{tabular}
\label{tab:baseline_config}
\end{table}

These baselines represent the dominant neural network paradigm identified in the literature review: end-to-end supervised models trained from scratch on labeled surface-to-target pairs with no pretraining or representation learning component.

\subsection{Masked Autoencoder Architecture}

The masked autoencoder consists of a symmetric encoder-decoder stack of stacked recurrent layers, followed by a TimeDistributed projection head that reconstructs the input sequence. The architecture is parameterized by three quantities: the number of encoder layers $L$, the recurrent cell type (LSTM or GRU), and the latent space width $d_z$.

\subsubsection*{Layer width schedule}

The encoder compresses the input representation progressively from the input feature dimension $F = 5$ down to the latent width $d_z = \lfloor F \cdot p_z \rceil$, where $p_z \in \{0.2, 0.5, 0.8\}$ is the latent percentage and $\lfloor \cdot \rceil$ denotes rounding to the nearest integer. For $L > 1$, the hidden widths at each encoder layer are spaced linearly between $F$ and $d_z$, then rounded to integers, giving an encoder width sequence that decreases monotonically from input to bottleneck. The decoder is the exact mirror: its layer widths are the encoder widths in reverse order. For $L = 1$, the encoder and decoder each contain a single layer of width $d_z$, placing two consecutive bottleneck layers at the center of the stack. The complete layer width sequence for the full autoencoder is therefore:

\begin{equation}
    \mathbf{w} =
    \underbrace{[w_1, \ldots, w_{L-1}]}_{\text{encoder hidden}}
    \;
    \underbrace{[d_z,\; d_z]}_{\text{bottleneck}}
    \;
    \underbrace{[w_{L-1}, \ldots, w_1]}_{\text{decoder hidden}}
\label{eq:widths}
\end{equation}

\noindent giving $2L$ recurrent layers in total. Every layer propagates the full sequence at each stage.

\subsubsection*{Output projection}

A \texttt{TimeDistributed(Dense($F$, activation=`linear'))} layer is applied after the final decoder layer to project back to the original $F$-dimensional feature space at every timestep, producing a reconstruction $\hat{\mathbf{X}} \in \mathbb{R}^{600 \times F}$ of the same shape as the input.

\subsubsection*{Random masking}

During pretraining, each input window $\mathbf{X}$ is corrupted by randomly zeroing a fraction $p_m$ of its elements before being fed to the encoder. The mask is applied element-wise across both the time and feature dimensions: $n_{\text{mask}} = \lfloor |\mathbf{X}| \cdot p_m \rfloor$ scalar positions are sampled uniformly without replacement from the full tensor and set to zero. The reconstruction target is always the \emph{unmasked} original $\mathbf{X}$, so the model must infer the missing values from the unmasked context. This masking scheme is applied on-the-fly inside the data generator, generating a fresh random mask for each window at every training step.

\subsubsection*{Training stability}

LSTM and GRU cells use their default internal activations (tanh for the recurrent state, sigmoid for the gates). The Adam optimizer is configured with gradient norm clipping to prevent gradient explosion. A \texttt{TerminateOnNaN} callback halts training immediately if any loss value becomes NaN, preventing cascading failures from propagating across epochs.

\subsection{Two-Stage Transfer Learning Pipeline}

Training proceeds in two sequential stages. The encoder weights learned in Stage~1 are carried forward and frozen in Stage~2; no parameters are shared between stages in any other way.

\subsubsection*{Stage 1: Self-supervised masked reconstruction}

The full symmetric autoencoder (encoder + decoder + output projection) is trained on the \emph{unlabeled} training windows, the input sequences with the target column removed, with no mud volume labels used. At each step, a fresh random mask is applied to the input batch, and the network is trained to reconstruct the original unmasked sequence. The loss is MAE computed over all $600 \times F$ output positions:

\begin{equation}
    \mathcal{L}_{\text{AE}} =
    \frac{1}{600 \cdot F}
    \sum_{t=1}^{600} \sum_{f=1}^{F}
    \left| X_{t,f} - \hat{X}_{t,f} \right|
\label{eq:ae_loss}
\end{equation}

Training runs for up to 10 epochs with the Adam optimizer, batch size 64, and early stopping with patience 5 monitoring validation reconstruction loss.

\subsubsection*{Stage 2: Supervised task header fine-tuning}

After pretraining, the decoder and output projection layers are discarded. The encoder half of the autoencoder, the first $L$ recurrent layers in $\mathbf{w}$, is extracted by counting the total number of recurrent layers in the trained model, taking the first half, and reconstructing them as a new functional model with their weights copied from Stage~1. The final encoder layer is reconfigured with $\mathtt{return\_sequences=False}$ so that it emits a single context vector rather than a full sequence. All extracted encoder layers are frozen, ensuring that pretraining representations are not degraded by the regression objective.

A trainable task header is appended to the frozen encoder output. The header consists of $L_h \in \{1, 2\}$ recurrent layers of the same cell type as the encoder, followed by a Dense(1) output unit with linear activation. The full model (frozen encoder + trainable header) is compiled with MAE loss and trained on the labeled $(\mathbf{X}, y)$ pairs for up to 15 epochs with early stopping (patience 5, monitoring validation MAE), batch size 64.

\subsection{Design Space Exploration for MAE Architectures}

No prior work establishes a standard MAE architecture for drilling telemetry, so a systematic full-factorial search over the MAE design space is conducted rather than committing to a single hand-tuned configuration. The search explores five dimensions, yielding 72 configurations in total. The dimensions and their candidate values are listed in Table~\ref{tab:search_space}, along with the rationale for each choice.

\begin{table}[!h]
\renewcommand{\arraystretch}{1.2}
\caption{MAE Design Space Search Dimensions}
\centering
\begin{tabular}{lll}
\textbf{Dimension} & \textbf{Values} & \textbf{Configs} \\
\hline
Encoder depth $L$           & \{1, 2\}              & 2 \\
\hline
Latent width $p_z$          & \{20\%, 50\%, 80\%\}  & 3 \\
\hline
Task header depth $L_h$     & \{1, 2\}              & 2 \\
\hline
RNN cell type               & \{LSTM, GRU\}         & 2 \\
\hline
Masking ratio $p_m$         & \{20\%, 50\%, 80\%\}  & 3 \\
\hline
\textbf{Total}              &                       & \textbf{72} \\
\hline
\end{tabular}
\label{tab:search_space}
\end{table}

The rationale for each dimension and its candidate values is as follows. \textbf{Encoder depth} $L \in \{1, 2\}$: shallow depths are explored to avoid over-parameterization on the 5-dimensional telemetry input, and both values represent tractable encoder configurations given the available pretraining data volume. \textbf{Latent width} $p_z \in \{20\%, 50\%, 80\%\}$: the three levels span a wide range of information compression, from aggressive bottlenecking (20\%, which rounds to 1 unit for $F=5$) to near-identity compression (80\%, 4 units). The three-level scheme follows the convention established in prior masked autoencoder studies~\cite{He_Chen_Xie_Li_Dollar_Girshick_2022, tang2022mtsmaemaskedautoencodersmultivariate}, which found that bottleneck width significantly affects downstream task performance. \textbf{Task header depth} $L_h \in \{1, 2\}$: two-layer headers have shown improved adaptation capacity in parameter-efficient transfer learning settings~\cite{houlsby2019parameterefficienttransferlearningnlp}, and both values are explored to determine empirically whether additional header capacity helps on this task. \textbf{RNN cell type} $\in \{$LSTM, GRU$\}$: both dominant recurrent cell types in the drilling analytics literature are evaluated, as prior work on time-series tasks has found task-dependent differences between them~\cite{ZhangCheng-Kai2023Bhpp}, motivating empirical comparison rather than selection by prior. \textbf{Masking ratio} $p_m \in \{20\%, 50\%, 80\%\}$: these three levels follow the masking ratio convention of the original MAE work~\cite{He_Chen_Xie_Li_Dollar_Girshick_2022}, which found that high masking ratios produced the best image representations; the three levels are explored here because the optimal ratio for low-dimensional time-series data may differ substantially from the image domain.

All other hyperparameters are held fixed across the search at the values used for the baselines: Adam optimizer, learning rate $0.001$, batch size 64, and MAE loss for both training stages. Early stopping with patience 5 is applied at both Stage~1 and Stage~2 of every configuration. The configuration achieving the lowest test MAE on the held-out test set after Stage~2 fine-tuning is selected as the best MAE model for final comparison against the baselines.

\subsection{Evaluation Protocol}

All models, both baselines and all 72 MAE configurations, are evaluated on the same held-out 20\% test set. No hyperparameter decisions are made on the basis of test set performance; the test set is accessed only once per model, after training is complete.

Two metrics are reported. Mean absolute error (MAE) is the primary metric, as it directly matches the training loss and is interpretable in the normalized target units:

\begin{equation}
    \text{MAE} = \frac{1}{N} \sum_{i=1}^{N} \left| y_i - \hat{y}_i \right|
\label{eq:mae}
\end{equation}

Root mean squared error (RMSE) is reported as a secondary metric, as its quadratic penalty gives larger weight to outlier predictions, providing a complementary view of model reliability:

\begin{equation}
    \text{RMSE} = \sqrt{\frac{1}{N} \sum_{i=1}^{N} \left( y_i - \hat{y}_i \right)^2}
\label{eq:rmse}
\end{equation}

\noindent where $y_i$ is the true window-mean Total Mud Volume and $\hat{y}_i$ is the model prediction for the $i$-th test window. All experiments are run with a fixed random seed to ensure reproducibility.

% ============================================================
\section{Results}
% ============================================================

\subsection{Baseline Model Performance}

The two supervised baselines establish the performance level against which all MAE configurations are compared. The LSTM baseline achieved a test MAE of $0.01959$ and the GRU baseline achieved a test MAE of $0.02599$, placing the LSTM lower in absolute error than the GRU. Both baselines use the same architecture, optimizer, and training procedure, differing only in cell type; the performance gap therefore reflects the difference between LSTM and GRU cells on this particular dataset and task.

\subsection{MAE Design Space Exploration Results}
\label{subsec:dse_results}

Across all 72 configurations, test MAE ranged from $0.02085$ (best) to $0.05034$ (worst), with a median of $0.02985$. None of the 72 MAE configurations outperformed the LSTM baseline. Of the 72 configurations, 9 (12.5\%) achieved lower MAE than the GRU baseline; the remaining 63 (87.5\%) were worse. The complete ranking of the top 10 configurations is shown in Table~\ref{tab:top10}.

\begin{table}[!h]
\renewcommand{\arraystretch}{1.15}
\caption{Top 10 MAE Configurations Ranked by Test MAE}
\centering
\small
\begin{tabular}{clcccc}
\textbf{Rank} & \textbf{Config (condensed)} & \textbf{MAE} & $L$ & $p_z$ & $p_m$ \\
\hline
1  & ae1-lat80-hd2-GRU-m20  & 0.02085 & 1 & 80\% & 20\% \\
2  & ae2-lat80-hd2-GRU-m20  & 0.02219 & 2 & 80\% & 20\% \\
3  & ae1-lat80-hd1-GRU-m20  & 0.02247 & 1 & 80\% & 20\% \\
4  & ae1-lat80-hd2-GRU-m50  & 0.02334 & 1 & 80\% & 50\% \\
5  & ae1-lat80-hd1-LSTM-m20 & 0.02446 & 1 & 80\% & 20\% \\
6  & ae1-lat80-hd2-GRU-m80  & 0.02516 & 1 & 80\% & 80\% \\
7  & ae1-lat80-hd2-LSTM-m20 & 0.02532 & 1 & 80\% & 20\% \\
8  & ae1-lat80-hd1-GRU-m80  & 0.02550 & 1 & 80\% & 20\% \\
9  & ae2-lat80-hd2-GRU-m50  & 0.02551 & 2 & 80\% & 50\% \\
10 & ae2-lat50-hd2-GRU-m50  & 0.02622 & 2 & 50\% & 50\% \\
\hline
\end{tabular}
\label{tab:top10}
\end{table}

\noindent ($L$ = encoder depth, $p_z$ = latent width percentage,
$p_m$ = masking ratio; all configurations use Adam, lr=$0.001$,
batch size 64.)

\subsection{Best MAE Configuration vs.\ Baselines}
\label{subsec:best_vs_baseline}

The best-performing MAE configuration used a single-layer encoder ($L=1$), a wide latent space ($p_z = 80\%$), a two-layer task header ($L_h = 2$), GRU cells, and a 20\% masking ratio. It achieved a test MAE of $0.02085$, which is $6.4\%$ higher than the LSTM baseline ($\Delta = +0.00126$) and $19.8\%$ lower than the GRU baseline ($\Delta = -0.00514$). Table~\ref{tab:comparison} presents the direct comparison.

\begin{table}[!h]
\renewcommand{\arraystretch}{1.2}
\caption{Best MAE Configuration vs.\ Supervised Baselines}
\centering
\begin{tabular}{lccc}
\textbf{Model} & \textbf{Test MAE} & \textbf{vs LSTM} & \textbf{vs GRU} \\
\hline
LSTM baseline          & 0.01959 & ,          & $-24.6\%$ \\
GRU baseline           & 0.02599 & $+32.7\%$   & ,        \\
Best MAE (GRU, lat80)  & 0.02085 & $+6.4\%$    & $-19.8\%$ \\
\hline
\end{tabular}
\label{tab:comparison}
\end{table}

The best MAE configuration therefore occupies an intermediate position between the two baselines: it substantially improves over the supervised GRU but does not yet close the gap to the supervised LSTM.

\subsection{Effect of Individual Design Dimensions}
\label{subsec:dimensions}

\subsubsection*{Latent space width}

\begin{figure}[htbp]
\centerline{\includegraphics[width=0.42\textwidth]{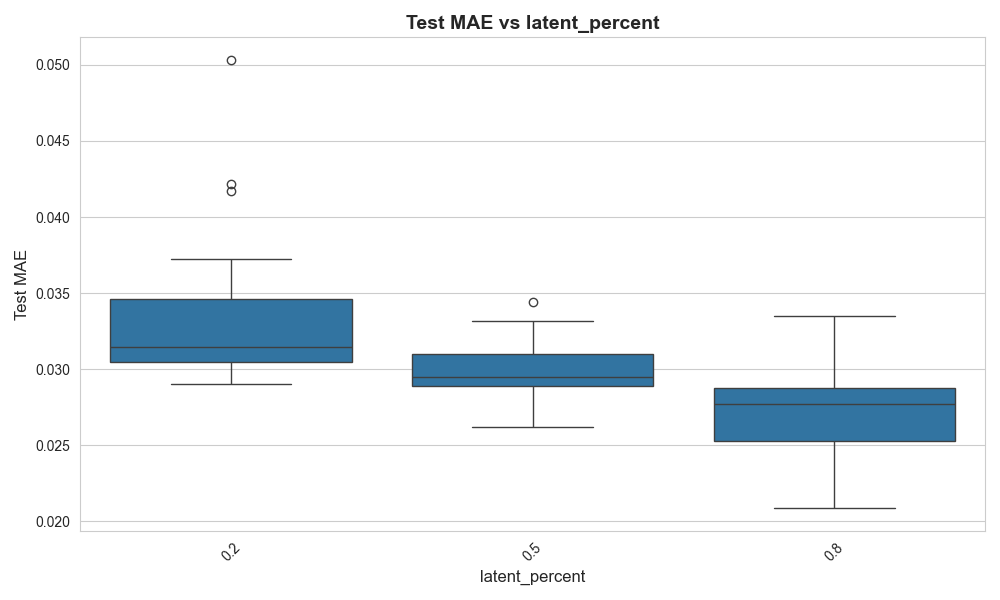}}
\caption{Test MAE distributions as a function of latent space width $p_z \in \{20\%, 50\%, 80\%\}$ across all 72 MAE configurations. Median MAE decreases monotonically from $0.03149$ at $p_z = 20\%$ to $0.02773$ at $p_z = 80\%$. All 9 configurations that outperformed the supervised GRU baseline used $p_z = 80\%$; no configuration with $p_z \leq 50\%$ achieved the same. Readers should note the shrinking IQR as $p_z$ increases, indicating that a wider bottleneck also stabilizes performance across other design choices. The key takeaway is that aggressive information compression (as low as 1 hidden unit at $p_z = 20\%$ for $F=5$) destroys too much temporal structure for the downstream regression task to recover, making latent width the single most important architectural decision for MAE pretraining on drilling telemetry.}
\label{fig:boxplot_latent}
\end{figure}

\begin{figure*}[htbp]
\centerline{\includegraphics[width=0.94\textwidth]{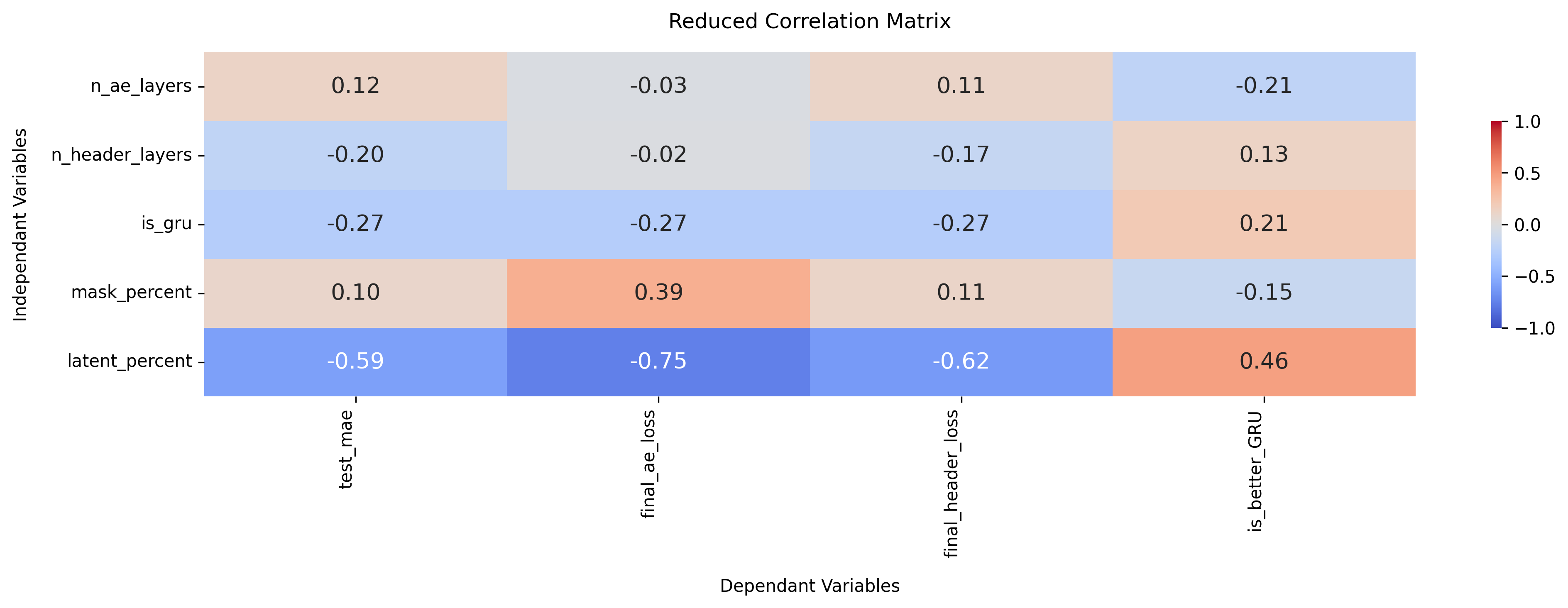}}
\caption{Reduced Pearson correlation matrix between the five MAE design dimensions (rows) and four performance metrics (columns). Cell color encodes correlation strength: deep blue is strong negative, deep red is strong positive. The dominant finding is the bottom row: \texttt{latent\_percent} ($p_z$) correlates strongly and negatively with all three loss metrics ($r \leq -0.59$), confirming it as the single most important design choice. Readers should also note that \texttt{mask\_percent} correlates only weakly with test MAE ($r = 0.10$), indicating masking ratio is largely inconsequential for this dataset. The rightmost column (\texttt{is\_better\_GRU}) flips sign relative to the loss columns, as expected: dimensions that reduce MAE also increase the probability of outperforming the GRU baseline.}
\label{fig:corr_matrix}
\end{figure*}

Latent space width is the dominant design dimension. As shown in Fig.~\ref{fig:boxplot_latent}, median MAE decreases monotonically with increasing $p_z$: configurations with $p_z = 20\%$ achieved a median MAE of $0.03149$, falling to $0.02950$ at $p_z = 50\%$ and $0.02773$ at $p_z = 80\%$. All 9 configurations that outperformed the GRU baseline used $p_z = 80\%$, and no configuration with $p_z \leq 50\%$ beat the GRU baseline. This monotonic trend is also confirmed by the reduced correlation matrix (Fig.~\ref{fig:corr_matrix}), which shows that $p_z$ has a Pearson correlation of $r = -0.59$ with test MAE, the strongest correlation of any design dimension with any performance metric in the search.

\subsubsection*{RNN cell type}

\begin{figure}[htbp]
\centerline{\includegraphics[width=0.42\textwidth]{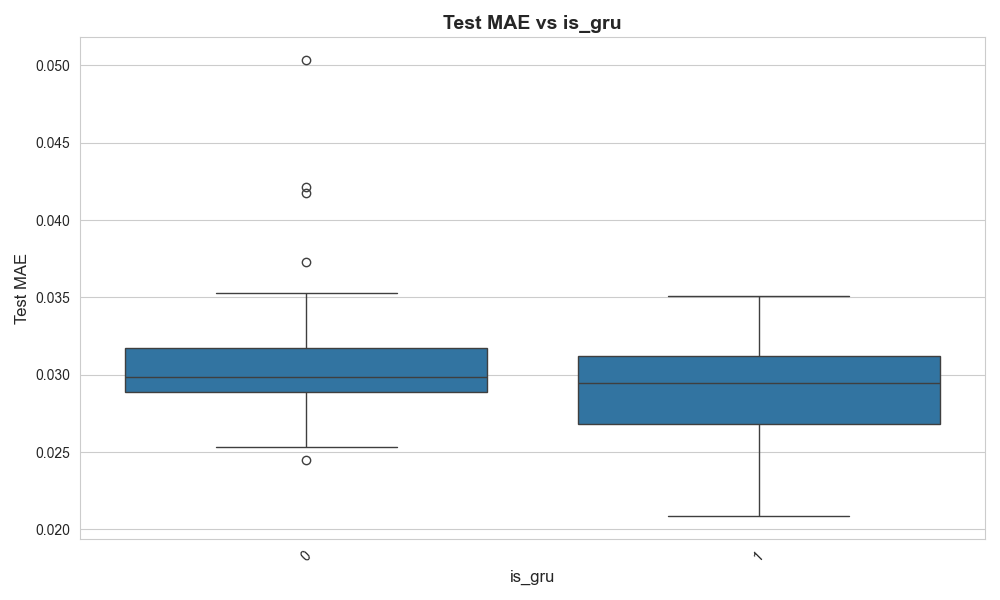}}
\caption{Test MAE distributions for LSTM-based ($\mathtt{is\_gru}=0$) versus GRU-based ($\mathtt{is\_gru}=1$) MAE configurations across all 72 experiments. Each box spans the interquartile range; the horizontal line is the median; whiskers extend to 1.5$\times$IQR; circles are outliers. GRU cells produce a lower median MAE ($0.02949$ vs.\ $0.02985$) and a tighter upper tail, indicating more consistent performance. The key takeaway is that GRU cells are the preferred cell type for MAE pretraining on this dataset, even though the fully supervised LSTM baseline outperforms the supervised GRU baseline, a reversal that suggests LSTM's additional gating complexity hinders representation learning under the reconstruction objective with limited pretraining data.}
\label{fig:boxplot_gru}
\end{figure}

GRU-based MAE configurations consistently outperformed LSTM-based configurations. As shown in Fig.~\ref{fig:boxplot_gru}, GRU configurations achieved a median MAE of $0.02949$ versus $0.02985$ for LSTM configurations. All 9 configurations that beat the GRU baseline were either GRU-celled or, in two cases (ranks 5 and 7), LSTM-celled with the widest latent space. The correlation matrix confirms this: $\mathtt{is\_gru}$ has a correlation of $r = -0.27$ with test MAE, the second strongest single-dimension association after latent percent. This result is notable because the supervised LSTM baseline outperforms the supervised GRU baseline, revealing that cell-type advantage does not transfer across training paradigms.

\subsubsection*{Masking ratio}

\begin{figure}[htbp]
\centerline{\includegraphics[width=0.42\textwidth]{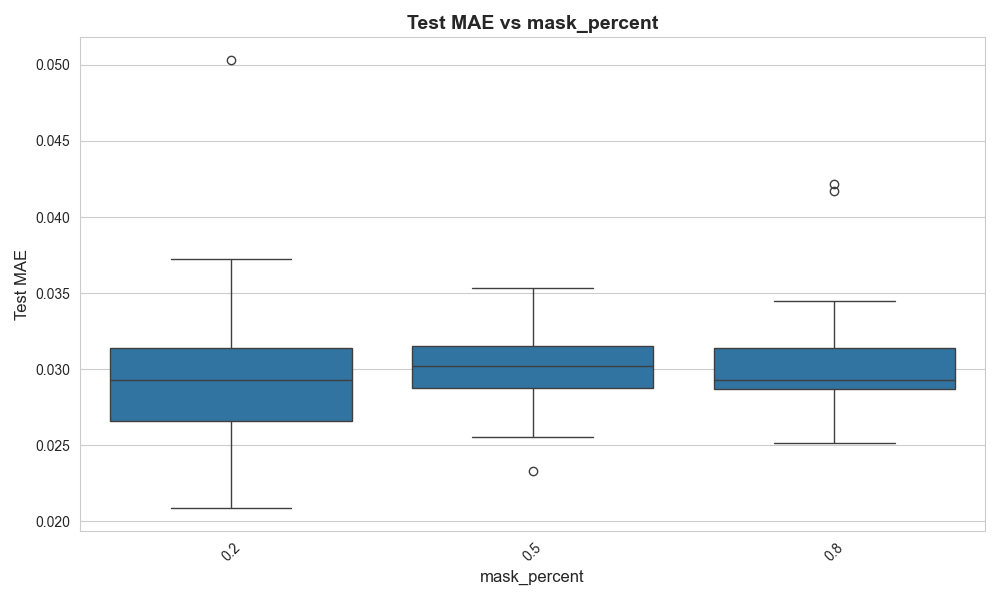}}
\caption{Test MAE distributions as a function of masking ratio $p_m \in \{20\%, 50\%, 80\%\}$ across all 72 MAE configurations. The three distributions are nearly identical in median ($0.02929$, $0.03024$, $0.02927$) and spread, indicating masking ratio has negligible influence on downstream prediction error. Readers should note the absence of any monotonic trend across the three levels, a sharp contrast with findings in image MAEs, where higher masking ratios consistently improve representations. The key takeaway is that temporal redundancy in 1~Hz drilling telemetry renders masking ratio a non-critical design choice, and practitioners should prioritize latent space width over masking schedule when configuring MAE pretraining for drilling data.}
\label{fig:boxplot_mask}
\end{figure}

Masking ratio has negligible effect on test MAE. As shown in Fig.~\ref{fig:boxplot_mask}, median MAEs for $p_m = 20\%$, $50\%$, and $80\%$ are $0.02929$, $0.03024$, and $0.02927$ respectively, a spread of less than $0.001$ across all three levels. The correlation matrix corroborates this: $p_m$ has a correlation of only $r = 0.10$ with test MAE. The best individual configuration used $p_m = 20\%$, but three of the top-10 configurations (ranks 4, 9, and 10) used $p_m = 50\%$, indicating that low masking is not a reliable prerequisite for good performance.

\subsubsection*{Encoder depth}

\begin{figure}[htbp]
\centerline{\includegraphics[width=0.42\textwidth]{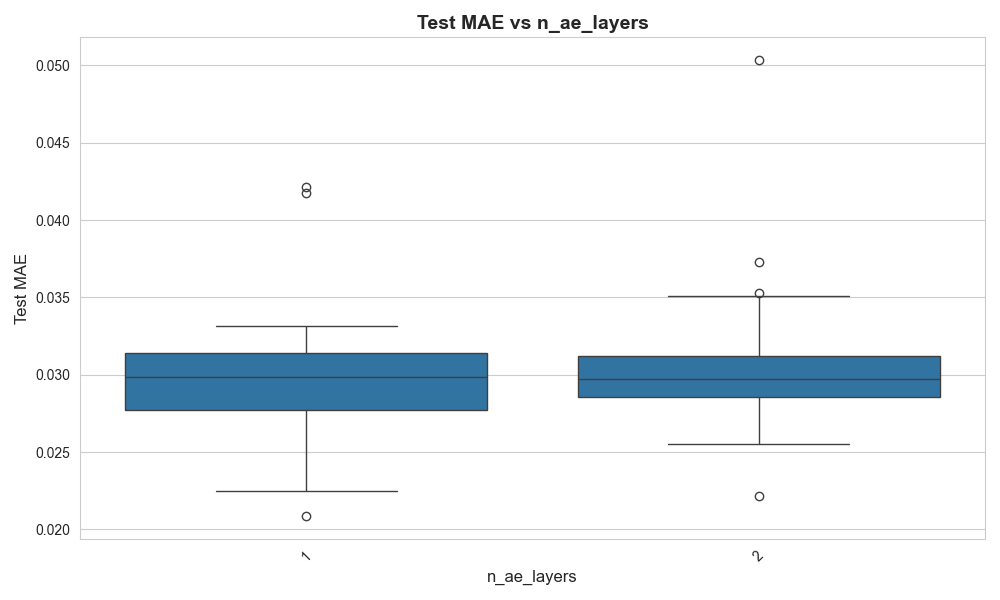}}
\caption{Test MAE distributions for single-layer ($L=1$) versus two-layer ($L=2$) encoders across all 72 MAE configurations. Median MAEs are $0.02986$ and $0.02975$ respectively, a difference of only $0.0001$. The near-identical distributions confirm that encoder depth within the $\{1, 2\}$ range explored here is not a meaningful axis of variation. The key takeaway is that for low-dimensional 5-channel drilling telemetry, a single recurrent encoder layer captures as much pretraining signal as two layers, and adding depth does not provide a representational benefit under the current experimental conditions.}
\label{fig:boxplot_ae_layers}
\end{figure}

Encoder depth has minimal impact on performance. One-layer and two-layer encoders produce median MAEs of $0.02986$ and $0.02975$ respectively, a difference of $0.0001$. The correlation matrix assigns $r = 0.12$ between encoder depth and test MAE. As shown in Fig.~\ref{fig:boxplot_ae_layers}, the interquartile ranges and whisker extents are nearly identical for both encoder depths, confirming that depth is not a meaningful axis of variation within the $\{1, 2\}$ range explored here.

\subsubsection*{Task header depth}

\begin{figure}[htbp]
\centerline{\includegraphics[width=0.42\textwidth]{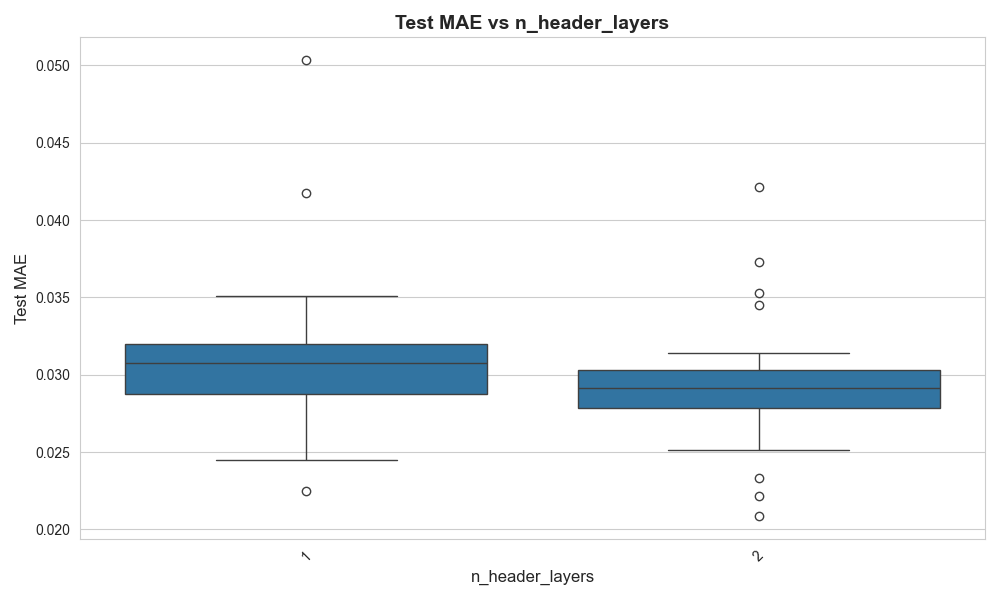}}
\caption{Test MAE distributions for one-layer ($L_h=1$) versus two-layer ($L_h=2$) task headers across all 72 MAE configurations. The two-layer header reduces the median MAE from $0.03076$ to $0.02915$ (a $5.2\%$ improvement) and achieves a lower minimum, indicating that additional adaptation capacity over the frozen encoder is consistently beneficial. The key takeaway is that practitioners should prefer a two-layer task header when deploying MAE pretraining for downhole regression, as the extra parameters provide meaningful adaptation to the regression objective without being trained from scratch.}
\label{fig:boxplot_header_layers}
\end{figure}

A two-layer task header provides a small but consistent advantage. Median MAE is $0.02915$ for $L_h = 2$ versus $0.03076$ for $L_h = 1$, a reduction of approximately $5.2\%$ in the median. The correlation matrix shows $r = -0.20$ for task header depth with test MAE. As shown in Fig.~\ref{fig:boxplot_header_layers}, the two-layer header distribution is shifted downward and has a notably lower minimum, suggesting the additional header capacity helps the model adapt the frozen representations to the regression task.

% ============================================================
\section{Discussion}
\label{sec:discussion}
% ============================================================

\subsection{Does MAE Pretraining Help?}
\label{subsec:does_it_help}

The answer depends critically on which baseline is used for comparison. Against the GRU baseline, MAE pretraining offers a clear advantage: the best MAE configuration reduces test MAE by $19.8\%$ relative to a supervised GRU trained on the same data. More broadly, 9 of 72 MAE configurations beat the GRU baseline, and all 9 do so by applying pretraining on top of a GRU encoder, suggesting that MAE pretraining is a consistently beneficial prior for the GRU cell type in this setting.

Against the LSTM baseline, MAE pretraining does not close the gap. The best MAE configuration is $6.4\%$ worse than the supervised LSTM, and no configuration in the entire design space achieves parity with it. The LSTM baseline therefore remains the stronger model for this task under the current experimental conditions. The data efficiency advantage of MAE pretraining, the core motivation for self-supervised learning, does not materialize within the scope of this study. However, the 19.8 \% MAE reduction achieved over the supervised GRU baseline illustrates the potential of self-supervised pretraining to accelerate digital transformation in drilling, offering operators a scalable route to more reliable downhole awareness in data-scarce geothermal and deep unconventional environments.

One likely explanation is the scale of the pretraining data. All experiments used a 20\% data subset for computational tractability, and self-supervised pretraining is generally known to yield diminishing benefits at small data scales. The pretraining signal from 600-timestep windows of 5-dimensional drilling telemetry may not be rich enough to provide a meaningful representational advantage over a supervised model trained from scratch on the same data. This interpretation is consistent with the broader MAE literature, in which large performance gains are typically reported at scale.

\subsection{Which Design Choices Matter Most?}
\label{subsec:design_choices}

The design space exploration reveals a clear hierarchy of importance among the five dimensions searched. Latent space width is by far the most consequential choice, with a Pearson correlation of $r = -0.59$ with test MAE across all 72 configurations. The monotonic improvement in median MAE as $p_z$ increases from $20\%$ to $50\%$ to $80\%$ suggests that the bottleneck in this architecture is genuinely harmful: compressing 5-dimensional drilling telemetry to as little as 1 unit ($p_z = 20\%$ rounds to $\lfloor 5 \times 0.2 \rceil = 1$) destroys too much information for the downstream regression task to recover.

RNN cell type is the second most important dimension ($r = -0.27$), with GRU cells consistently outperforming LSTM cells in the MAE setting. This is counterintuitive given that the supervised LSTM baseline outperforms the supervised GRU baseline. The asymmetry suggests that LSTM's additional gating complexity makes the encoder harder to train to a useful representation under the reconstruction objective with limited data and only 10 pretraining epochs.

Masking ratio, encoder depth, and task header depth all show weak correlations with test MAE ($|r| \leq 0.20$). The insensitivity to masking ratio is the most surprising finding. The most plausible explanation is that 5-dimensional drilling telemetry is highly temporally redundant at 1~Hz: adjacent timesteps carry almost identical information, so whether 20\% or 80\% of elements are zeroed, the model can reconstruct the masked values from nearby context with similar ease. This removes the usual pressure that high masking ratios exert on the encoder to learn long-range dependencies, collapsing the distinction between masking levels.

\subsection{Why Self-Supervised Pretraining Is Partially Effective}
\label{subsec:why_effective}

Despite not outperforming the LSTM baseline, the MAE pretraining approach does learn useful representations for Total Mud Volume prediction. The systematic advantage of GRU-based MAE over the supervised GRU baseline demonstrates that pretraining adds signal beyond what supervised training alone can achieve. The physical properties of drilling telemetry make it a promising domain for masked reconstruction pretraining: mud volume, depth, pump output, and weight-on-bit are strongly coupled over time through the mechanics of the drilling system. Masking a fraction of these channels forces the encoder to learn the inter-variable dependencies that underlie these couplings rather than simply copying inputs to outputs. The resulting latent representations appear to encode aspects of the drilling state that are predictive of mud volume, even when learned without mud volume labels.

% ============================================================
\section{Limitations}
\label{sec:limitations}
% ============================================================

Five key limitations of this study should be considered when interpreting the results.

\textbf{Data subset.} All experiments used a 20\% subset of the available data for computational tractability. Self-supervised pretraining is known to be particularly sensitive to pretraining corpus size, with performance gains typically increasing as pretraining data scales. The relative performance of MAE configurations versus the LSTM baseline may improve at full data scale, and the current results may underestimate the potential of the MAE paradigm.

\textbf{Two-well scope.} Both wells were drilled in the same Utah FORGE geothermal formation with similar rig configurations. The results therefore reflect performance on a single geological context, and generalization to different formations, fluid systems, or drilling environments has not been evaluated. The absence of cross-well or cross-formation validation means that it is not possible to assess whether the encoder representations learned during pretraining transfer across wells with different geological characteristics.

\textbf{Fixed auxiliary hyperparameters.} The design space search held several potentially important hyperparameters fixed: learning rate, batch size, and number of pretraining epochs were not varied. These hyperparameters may interact with the architectural choices explored here, for example, wider latent spaces or higher masking ratios may respond differently to learning rate schedule changes, and the fixed hyperparameter assumption may have disadvantaged certain configurations in the search.

\textbf{Frozen encoder.} The encoder is fully frozen during Stage~2 fine-tuning. While freezing prevents the regression objective from degrading the pretraining representations, it also prevents the encoder from adapting to task-specific structure that may not have been captured during pretraining. Partial unfreezing or parameter-efficient fine-tuning strategies such as low-rank adaptation (LoRA)~\cite{houlsby2019parameterefficienttransferlearningnlp} may allow the encoder to close the remaining gap to the supervised LSTM baseline without losing the benefits of pretraining.

\textbf{Scalar target only.} The prediction target is the window-mean of Total Mud Volume, a scalar summary statistic. This formulation may underutilize the temporal encoder representations produced by the recurrent stack, which encode the full sequence dynamics. Sequence-level prediction targets or multi-step forecasting formulations may better exploit the encoder's representational capacity and produce larger gains over purely supervised approaches.

% ============================================================
\section{Future Work}
\label{sec:future_work}
% ============================================================

Several directions emerge directly from the limitations and findings of this study.

\textbf{Cross-well transfer.} The core value proposition of MAE pretraining is data efficiency: a model pretrained on one well should require fewer labeled examples from a second well to achieve strong performance. Empirically evaluating this by pretraining on one FORGE well and fine-tuning with progressively smaller labeled subsets from the second well would directly test whether MAE pretraining provides the sample efficiency gains that motivate self-supervised learning.

\textbf{Scaling study.} Systematically varying the pretraining data volume from 5\% to 100\% of the available dataset and measuring the effect on both Stage~1 reconstruction loss and Stage~2 downstream MAE would identify the minimum data requirements for MAE benefit and characterize the scaling behavior of the pretraining-to-fine-tuning transfer.

\textbf{Partial encoder unfreezing and LoRA.} Exploring partial encoder unfreezing and low-rank adaptation (LoRA) strategies in Stage~2 would determine whether allowing limited encoder adaptation narrows the gap to the supervised LSTM baseline. A careful ablation comparing fully frozen, partially frozen, and fully unfrozen encoders, with and without LoRA, would provide guidance for practitioners deploying MAE pretraining on small labeled datasets.

\textbf{Multi-task fine-tuning.} A single frozen encoder pretrained on surface telemetry could be used simultaneously to predict ECD, BHP, and downhole string vibrations by appending separate task headers for each target. Shared encoder representations across multiple downhole targets may improve per-task performance and reduce the total labeled data required across all prediction objectives.

\textbf{Anomaly detection via reconstruction error.} The frozen MAE encoder's reconstruction error provides a natural anomaly score that requires no labeled anomaly examples. Applying this to drilling event detection, including kicks, lost circulation, and pack-off events, would evaluate whether the encoder's learned representations capture the statistical regularities of normal drilling operation well enough to flag deviations in an unsupervised setting.

% ============================================================
\section{Conclusion}
\label{sec:conclusion}
% ============================================================

In this work, we presented the first empirical evaluation of masked autoencoder (MAE) pretraining for downhole drilling metric prediction. Motivated by our systematic review of 13 drilling analytics papers that identified 21 distinct supervised methods but zero applications of self-supervised or autoencoder-based pretraining, we designed a two-stage transfer learning pipeline and evaluated it against supervised LSTM and GRU baselines on real Utah FORGE geothermal well drilling telemetry.

A full-factorial design space search across 72 MAE configurations, spanning encoder depth, latent space width, masking ratio, RNN cell type, and task header depth, revealed that the best configuration reduces test MAE by 19.8\% relative to the supervised GRU baseline while trailing the supervised LSTM baseline by 6.4\%. Analysis of design dimensions identified latent space width as the dominant architectural choice ($r = -0.59$ with test MAE): all 9 configurations that outperformed the GRU baseline used the widest latent space ($p_z = 80\%$), and no configuration with a narrower bottleneck achieved the same. Masking ratio, contrary to expectations from the image MAE literature, had negligible effect ($r = 0.10$), a finding attributed to the high temporal redundancy of 1~Hz drilling telemetry.

Together, these results establish three conclusions. First, MAE pretraining is a viable paradigm for drilling analytics: self-supervised representations learned from unlabeled surface telemetry transfer meaningfully to downhole metric prediction. Second, the benefit of MAE pretraining is architecture-dependent and data-scale-dependent: a wide latent space and GRU cells are prerequisites for competitive performance under the current experimental conditions, and the data efficiency advantage over fully supervised training may widen at larger pretraining scales. Third, the negligible effect of masking ratio highlights a domain-specific challenge, temporal redundancy in drilling telemetry, that must be addressed in future architectures, for example through block masking or downsampled masking strategies designed to force long-range temporal reasoning.

As geothermal and oil-and-gas drilling operations increasingly generate high-frequency surface telemetry, self-supervised pretraining offers a principled path toward more data-efficient downhole prediction models. Our work provides the first empirical foundation for that direction to advance sustainable subsurface energy resources through digitalization and artificial intelligence.

% ============================================================
\section*{References}
% ============================================================
\renewcommand{\refname}{}
\bibliographystyle{IEEEtran}
\bibliography{LiteratureReview}

\end{document}